%% file: main.tex
\documentclass{article}

\usepackage{arxiv}
\usepackage{defs}

\usepackage[utf8]{inputenc} 
\usepackage[T1]{fontenc}    
\usepackage{hyperref}       
\usepackage{url}            
\usepackage{booktabs}       
\usepackage{amsfonts}       
\usepackage{nicefrac}       
\usepackage{microtype}      
\usepackage{lipsum}
\usepackage{graphicx}
\usepackage{comment}
\usepackage{natbib}
\usepackage{tikz}
\usetikzlibrary{shadows,arrows.meta,positioning,backgrounds,fit,chains,scopes}
\tikzstyle{decision} = [diamond, draw, fill=blue!20, 
    text width=4.5em, text badly centered, node distance=3cm, inner sep=0pt]
\tikzstyle{io} = [draw, trapezium, trapezium left angle=70, trapezium right angle=110, 
    minimum width=3em, minimum height=2em, text badly centered, draw=black, fill=blue!20]
\tikzstyle{block} = [rectangle, draw, fill=blue!20, 
    text width=6em, text centered, rounded corners, minimum height=4em]
\tikzstyle{line} = [draw, -latex']
\tikzstyle{cloud} = [draw, ellipse,fill=red!20, node distance=3cm,
    minimum height=2em]

\newcommand{\corp}[0]{\ensuremath \mathcal{W}}
\newcommand{\old}[1]{\ensuremath #1^{\text{(old)}}}
\newcommand{\new}[1]{\ensuremath #1^{\text{(new)}}}
\renewcommand{\vec}[1]{\ensuremath \mathbf{#1}}
\newcommand{\embin}[0]{\ensuremath \vec{u}}
\newcommand{\Embin}[0]{\ensuremath \vec{U}}
\newcommand{\embout}[0]{\ensuremath \vec{v}}
\newcommand{\Embout}[0]{\ensuremath \vec{V}}
\renewcommand{\example}[1]{\textit{#1}}
\graphicspath{{./figs/}}

\title{Follow the Leader: Documents on the Leading Edge of Semantic Change Get More Citations\thanks{To appear in the Journal of the Association for Information Sciences and Technology}}

\author{
  Sandeep Soni \\
  School of Interactive Computing\\
  Georgia Institute of Technology\\
  \texttt{sandeepsoni@gatech.edu} \\
  \And
  Kristina Lerman \\
  Information Sciences Institute\\
  University of Southern California\\
  \texttt{lerman@isi.edu} \\
  \And
  Jacob Eisenstein \\
  Google Research \\
  \texttt{me@jacob-eisenstein.com}
}

\begin{document}
\maketitle

\input{abstract}


\input{introduction}
\input{model}
\input{data}
\input{experiments}
\input{results}
\input{testing}
\input{related}
\input{discussion}
\newpage
\input{supplementary}
\clearpage

\bibliographystyle{plainnat}  
\bibliography{progressiveness}

\end{document}

%% file: abstract.tex
\begin{abstract}
Diachronic word embeddings -- vector representations of words over time -- offer remarkable insights into the evolution of language and provide a tool for quantifying sociocultural change from text documents. Prior work has used such embeddings to identify shifts in the meaning of individual words. However, simply knowing that a word has changed in meaning is insufficient to identify the instances of word usage that convey the historical or the newer meaning. In this paper, we link diachronic word embeddings to documents, by situating those documents as leaders or laggards with respect to ongoing semantic changes. Specifically, we propose a novel method to quantify the degree of semantic progressiveness in each word usage, and then show how these usages can be aggregated to obtain scores for each document. We analyze two large collections of documents, representing legal opinions and scientific articles. Documents that are scored as semantically progressive receive a larger number of citations, indicating that they are especially influential. Our work thus provides a new technique for identifying lexical semantic leaders and demonstrates a new link between progressive use of language and influence in a citation network.    
\end{abstract}

%% file: introduction.tex
\section{Introduction}
\label{sec:introduction}
Languages are continuously evolving~\citep{weinreich1968empirical}, and one particularly salient aspects of language change is how elements such as words are repurposed to new meanings~\citep{traugott2001regularity}.
Word embeddings -- representations of words as vectors in high-dimensional spaces -- can identify semantic changes in text documents by tracking shifts in each word's distributional neighborhood~\citep{kutuzov2018diachronic}.
However, these methods treat each word in isolation and do not indicate where change takes place: which documents or passages introduce new meanings for words,
and which lag behind in adopting semantic changes?

The ability to identify documents in the vanguard of linguistic change would 
yield valuable insights into the life cycle of new ideas:
for example, by making it possible to identify and  
support innovation in science~\citep{fortunato2018science}, and would provide new evidence about the social processes underlying linguistic and scholarly influence~\citep{gerow2018measuring}.
As a step towards this goal, we propose a simple quantitative technique for identifying the leading examples of ongoing semantic changes.
Our method builds directly on the embedding-based techniques for detecting changes in large corpus of documents, and takes the form of a likelihood ratio comparison between ``older'' and ``newer'' embedding models.
Usages that are better aligned with the newer embedding model can be considered to be more semantically ``progressive,'' in the sense of reflecting newer word meanings.

Using large datasets of legal opinions and scientific research abstracts produced over a long period of time, we demonstrate that more semantically advanced usages are indeed associated with documents that are landmarks in their respective fields, such as prominent Supreme Court rulings and foundational research papers. We further formalize these insights by demonstrating a novel relationship between semantic progressiveness and citation counts: in both domains, semantically progressive documents receive more citations, even after controlling for document content and a range of structural factors. While previous work has identified connections between word frequency and impact, we are the first to link semantic changes to citation networks. To summarize the contributions of this paper:
\begin{itemize}
\item We identify markers of semantic change in scientific articles and legal opinions (both in English). Legal opinions have not previously been analyzed with respect to dynamic word embeddings, and have received little attention in natural language processing.
\item We propose a novel method to score documents on their semantic progressiveness, thereby identifying documents on the vanguard of semantic change.
\item We show that documents at the vanguard of semantic change tend to be more influential in citation networks. 
\end{itemize}

Diachronic word embeddings help in identifying the points of transition in the language, especially as it pertains to semantic changes. Past research in information science, especially around the science of science, has identified the importance of transitions in the scholarly process -- for example, in determining the impact of scientific works~\citep{gerow2018measuring}, understanding the impact of individual topics~\citep{yan2015research}, studying the formation of interdisciplinary research areas~\citep{xu2018understanding}, and in mapping out the structure of scientific communities over time~\citep{boyack2014creation}. As a result, we see this work as relevant for a broader objective of understanding the scholarly process through the transitional aspects of language use and its link to scholarly outcomes. Our proposed approach provides a quantitative way to identify semantically innovative documents and study their dynamics. A secondary contribution to information science is to provide a new link between two orthogonal perspectives on document collections: content analysis and citation structures. While the topic modeling literature has offered models that link citations to sets of co-occurring words~\citep[e.g.,][]{nallapati2008joint,ding2011scientific}, we offer a more fine-grained lexical semantic perspective by connecting incoming citations to leadership on changes in the meanings of individual words, showcasing the potential of diachronic word embeddings as a tool for information science.\footnote{We have released the code and the word embeddings for both the datasets at \url{https://github.com/sandeepsoni/semantic-progressiveness/}}

%% file: model.tex
\section{Measuring Semantic Progressiveness}
\label{sec:model}

\emph{Diachronic word embeddings} make it possible to measure lexical semantic change over time~\citep[e.g.,][]{kulkarni2015statistically,hamilton2016diachronic}. In standard word embeddings, each word type is associated with a vector of real numbers, based on its distributional statistics~\citep{turney2010frequency,mikolov2013distributed}. In diachronic word embeddings, this vector is time-dependent, reflecting how a word's meaning (and associated distributional statistics) can change over time. Building on diachronic word embeddings, our method is comprised of four steps: (1) learning diachronic embeddings of words; (2) identifying semantic innovations using their diachronic embeddings; (3) scoring each usage by its position with respect to the semantic change; and (4) aggregating these scores by document. A schematic of the entire pipeline is shown in~\autoref{fig:flowchart}. We now describe each of these steps in detail.

\subsection{Estimating Word Embeddings}
\label{subsec:diachronic}
Several methods to learn diachronic word embeddings have been proposed~\citep[e.g.,][]{bamler2017dynamic,frermann2016bayesian,hamilton2016diachronic,rosenfeld2018deep}. In this work, we use the method proposed by \cite{hamilton2016diachronic} as it is conceptually straightforward and offers flexibility in the choice of the embedding algorithm. The core of this approach is to fit embedding models to distinct time-slices of the corpora, and then align the resulting embeddings.

Formally, assume a finite vocabulary $\Vcal$, and two corpora, $\old{\corp}$ and $\new{\corp}$, where each corpus is a set of sequences of tokens, $\corp = \{(w_{i,1}, w_{i,2}, \ldots, w_{i, T_i})\}_{i=1}^N$, where $N$ is the number of documents in the corpus, $i$ indexes an individual document whose length is $T_i$, and each $w_{i,t} \in \Vcal$. For each corpus, we estimate a set of word embeddings on the single vocabulary $\Vcal$. Following \cite{hamilton2016diachronic}, we estimate skipgram embeddings~\citep{mikolov2013distributed}, which are based on the objective of predicting context words $w_{t'}$ conditioned on a target word $w_t$.

While the mathematical details of skipgram word embeddings are well known, they are crucial to our method for situating individual usages of words with respect to ongoing semantic changes.
For this reason, we present a brief review.
Omitting the document index $i$, the skipgram objective is based on the probability,
\begin{equation}
P(w_{t'} \mid w_t) \propto \exp \left( \embout_{w_{t'}} \cdot \embin_{w_t} \right),
\label{eq:skipgram-prob}
\end{equation}
where $\embout_{w_{t'}}$ is the embedding of $w_{t'}$ when it is used as a context (also called as the ``output'' embedding), and $\embin_{w_t}$ is the embedding of $w_{t}$ when it is used as a target word (also called as the ``input'' embedding).

Normalizing the probability in \autoref{eq:skipgram-prob} requires summing over all possible $w_{t'}$, which is computationally expensive.
Typically the skipgram estimation problem is solved by negative sampling~\citep{mikolov2013distributed}, but this does not yield properly normalized probabilities.
We therefore turn instead to noise-contrastive estimation~\citep[NCE;][]{gutmann2010noise}, which makes it possible to estimate the probability in \autoref{eq:skipgram-prob} without computing the normalization term~\citep{mnih2013learning}.

Suppose that the observed data is augmented with a set of ``noise'' examples $\{(\tilde{w}, w_t)\}$, where each $\tilde{w}$ is sampled from a unigram noise distribution $P_n$.
Further assume that there are $k$ noise examples for every real example.
An alternative prediction task is to decide whether each example is from the real data $(D=1)$ or from the noise $(D=0)$.
The cross entropy for this task is,
\begin{equation}
  \begin{split}
    J = &{} \sum_t \log \Pr(D = 1 \mid w_t, w_{t'})\\
    &{} + \sum_{j=1}^k \log \Pr(D = 0 \mid w_t, \tilde{w}^{(j)}),
  \end{split}
  \label{eq:nce-objective}
\end{equation}
where each $\tilde{w}^{(j)}$ is drawn from $P_n$.

Now let us define the probability,
\begin{align}
    \Pr(D = 1 \mid w_t, w_{t'})
    = &{} \frac{P(w_{t'} \mid D = 1, w_t)\Pr(D=1)}{P(w_{t'} \mid D=1, w_t)\Pr(D=1) + P(w_{t'} \mid D=0)\Pr(D=0)}\\
    = &{} \frac{P(w_{t'} \mid w_t)}{P(w_{t'} \mid w_t) + k P_n(w_{t'})}\\
    = &{} \left(1 + k\frac{P_n(w_{t'})}{P(w_{t'} \mid w_t)}\right)^{-1}\\
=  &{} \sigma\left( \embout_{w_{t'}} \cdot \embin_{w_t} - Z(w_t) - \log (k P_n(w_{t'})) \right)\\
\approx  &{} \sigma\left( \embout_{w_{t'}} \cdot \embin_{w_t} - \log (k P_n(w_{t'}) ) \right),
\label{eq:final-nce-objective}
\end{align}
where $\sigma$ indicates the sigmoid function $\sigma(x) = (1 + \exp(-x))^{-1}$. The log-normalization term $Z(w_t) = \log \sum_{w'} \exp \embout_{w'} \cdot \embin_{w_t}$ can be dropped in \autoref{eq:final-nce-objective} because the NCE objective is approximately ``self-normalizing'' when $P_n$ has positive support over all ${w \in \Vcal}$~\citep{mnih2013learning}.
We then maximize \autoref{eq:nce-objective} by gradient ascent, which yields embeddings that are asymptotically equivalent to the optimizers of \autoref{eq:skipgram-prob}~\citep{gutmann2010noise}.
Noise-contrastive estimation is closely related to the negative sampling objective typically employed in skipgram word embeddings, but of the two, only NCE-based embeddings can be interpreted probabilistically~\citep{dyer2014notes}, as required by our approach.

The skipgram model is not identifiable: any permutation of the dimensions of the input and output embeddings will yield the same result so many parameterizations of the model are observationally equivalent. For this reason, embeddings that are learned independently across multiple corpora must be aligned before their similarity can be quantified. To reconcile the input embeddings between the corpora $\old{\corp}$ and $\new{\corp}$ and make them comparable across the two corpora, we follow \cite{hamilton2016diachronic} and apply the Procrustes method~\citep{gower2004procrustes} to identify an orthogonal projection $\vec{Q}$ that minimizes the Frobenius norm $||\vec{Q}\old{\Embin} - \new{\Embin}||_F,$ where $||\vec{X}||_F = \sqrt{\sum_{i,j} x_{i,j}^2}$. 

\paragraph{Sensitivity to initialization} One potential downside of NCE is that its embeddings depend on the random initialization, unlike deterministic techniques such as singular value decomposition~\citep{sagi2011tracing,levy2014neural}.
As a result, the list of near neighbors can change across multiple runs~\citep{hellrich-hahn:2016:COLING}.
Nonetheless, we chose NCE because the resulting embeddings outperformed alternatives on intrinsic word similarity benchmarks~\citep{luong2013morpho}. 
Our robustness checks indicated that the method identified similar sets of semantic innovations across multiple runs.

\subsection{Discovering Semantic Innovations}
\label{subsec:semantic-innovations}
After estimating the diachronic embeddings for each word, the next step is to identify semantic innovations: words that have shifted in meaning.
One possibility would be to directly measure differences between the embeddings $\old{\embin}$ and $\new{\embin}$, but this can be unreliable because the density of embedding space is not guaranteed to be uniform.
We therefore follow the local second-order approach proposed by \cite{hamilton2016cultural}.
First, for each word we form the union of the sets of a word's near-neighbors ($n=50$) in the ``old'' and ``new'' periods.
Next, we compute the similarity of the word's embedding to the embeddings for members of this set, for both the ``old'' and ``new'' embeddings.
This yields a pair of vectors of similarities, each reflecting the word's position in a local neighborhood.
The degree of change in a word's position is the distance between these two vectors.

\subsection{Situating Usages with Respect to Semantic Change}
\label{subsec:progressiveness}
Given a set of semantic innovations $\Scal \subset \Vcal$, our main methodological innovation is to situate usage with respect to semantic changes. 
Each usage of an innovation $w^* \in \Scal$ can be analyzed using the likelihood function underlying the skipgram objective,
and scored by the ratio of the log-likelihoods under the embedding models associated with $\old{\corp}$ and $\new{\corp}$.
Specifically, we compute the sum,
\begin{align}
r_{w^*,i} = &{} \sum_{t : w_{i,t} = w^*} \sum\limits_{\substack{j \geq -k\\j\leq k\\j \neq 0}} 
\log \frac{\new{P}(w_{i,t+j} \mid w^*)}{\old{P}(w_{i,t+j}\mid w^*)}.
\end{align}
The intuition behind the statistic is to predict the context of every appearance of the semantic innovation $w^*$ in the document $i$ using both the ``new'' and ``old'' meaning of $w^*$ and the surrounding context. These new and old meanings are obtained from the embedding models associated with $\new{\corp}$ and $\old{\corp}$ respectively. Note that the document $i$ need not necessarily be in either $\old{\corp}$ or $\new{\corp}$. Substituting the form of probability from \autoref{eq:skipgram-prob} and simplifying further, the log-likelihood ratio reduces to:
\begin{equation}
\label{eq:progressiveness}
\begin{split}
r_{w^*,i} =  
\sum_{t : w_{i,t} = w^*} \sum\limits_{\substack{j \geq -k\\j\leq k\\j \neq 0}} 
&{}
\new{\embout}_{w_{i,t+j}} \cdot \new{\embin}_{w^*} - \new{Z}_{w^*}\\
&{} - \old{\embout}_{w_{i,t+j}} \cdot \old{\embin}_{w^*} + \old{Z}_{w^*}, 
\end{split}
\end{equation}
where $Z_{w^*}$ is the log normalization term, $\log \sum_{w'} \exp \left(\embout_{w'} \cdot \embin_{w^*}\right)$.
This metric intuitively favors documents that use $w^*$ in contexts that align with the new embeddings $\new{\embin}_{w^*}$ and $\new{\Embout}$.

\subsection{Aggregating to Document Scores}
\label{sec:doc-scores}
Given a set of innovations $\mathcal{S} \subset \mathcal{V}$, for each document $i$ we obtain a set of scores $\{r_{i,w^*} : w^* \in \mathcal{S}\}$.
The score for document $i$ is the maximum over the set of innovations, $m_i = \max_{w^* \in \mathcal{S}} r_{i,w^*}$.
This quantifies the maximal extent to which the document's lexical semantics match that of the more contemporary embedding model, $(\new{U}, \new{V})$.
We then standardize against other documents published in the same year, by computing the $z$-score, 
${z_i = \frac{m_i - \mu}{\sigma}},$
where $\mu$ is the mean score for documents published in the same year, and $\sigma$ is the standard deviation.
Documents with a positive $z$-score have lexical semantics that better match the contemporary embedding model than other documents written at the same time, and can thus be said to be semantically progressive. By standardizing each year separately, we ensure that the progressiveness metric does not inherently favor older or newer texts.

As a robustness check, we also experimented with an alternative discretized approach for scoring the documents. In this scheme, the document score was calculated as the number of innovations whose progressiveness exceeds the median progressiveness value over the entire set of innovations. The subsequent analysis of the documents with this scoring scheme produced qualitatively similar results to those obtained with the measure described in the previous paragraph (innovativeness of maximally innovative word per document), and so are only included in the supplementary material for this paper.

\input{flowchart}

%% file: flowchart.tex
\pagestyle{empty}

\tikzset{%
  materia/.style={draw, fill=blue!20, text width=12.0em, text centered, minimum height=1.5em,drop shadow},
  etape/.style={materia, text width=15em, minimum width=20em, minimum height=3em, rounded corners, drop shadow},
  linepart/.style={draw, thick, color=black!50, -LaTeX, dashed},
  line/.style={draw, thick, color=black!50, -LaTeX},
  ur/.style={draw, text centered, minimum height=0.01em},
  back group/.style={fill=yellow!20,rounded corners, draw=black!50, dashed, inner xsep=15pt, inner ysep=10pt},
}
\newcommand{\transreceptor}[3]{%
  \path [linepart] (#1.east) -- node [above] {\scriptsize #2} (#3);}

\begin{figure}
    \centering
    \begin{tikzpicture}
  [
    start chain=p going below,
    every on chain/.append style={etape},
    every join/.append style={line},
    node distance=1 and -.25,
  ]
  {
    \node [on chain, join] {Temporally bin the document collection};
    \node [on chain, join] {Learn word embeddings~\citep{mnih2013learning}};
    \node [on chain, join] {Align embeddings~\citep{hamilton2016diachronic}};
    \node [on chain, join] {Identify semantic innovations~\citep{hamilton2016cultural}};
    \node [on chain, join] {Calculate progressiveness per usage of every innovation};
    \node [on chain, join] {Calculate progressiveness per document};
  }

  \begin{scope}[on background layer]
    \node (bk3) [back group] [fit=(p-5) (p-6)] {};
    \node [draw, thick, green!50!black, fill=green!75!black!25, rounded corners, fit=(p-5) (p-6), inner xsep=15pt, inner ysep=10pt] {};
  \end{scope}

\end{tikzpicture}
    \caption{Flowchart shows our complete pipeline and highlights (in green) our methodological contributions.}
    \label{fig:flowchart}
\end{figure}
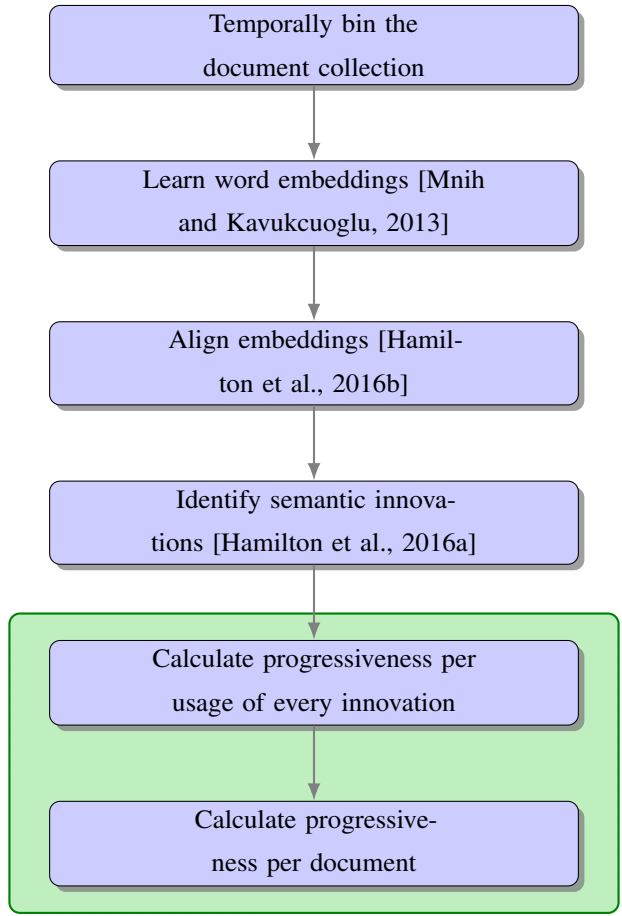

%% file: data.tex
\section{Data}
\label{sec:data}
We empirically validate 
our approach on two document collections: documents representing \textit{legal opinions} in federal courts of the United States of America~\citep{lerman2017bounded},\footnote{\url{https://www.courtlistener.com/}} and the DBLP collection of  \textit{computer science abstracts}~\citep{ley2002dblp}.\footnote{\url{https://dblp.uni-trier.de/}} These datasets were chosen because they include timestamps as well as citation information, making it possible to link semantic innovation with influence in a citation network.

\paragraph{Legal opinions.} A legal opinion is a document written by a judge or a judicial panel that summarizes their decision and all relevant facts about a court case.
We obtained all legal opinions by using the bulk API of a publicly available service.\footnote{\url{https://www.courtlistener.com/api/bulk-info/}}
These opinions span over $400$ courts, multiple centuries and have a broad jurisdictional coverage.

\paragraph{Scientific abstracts.} The abstracts from DBLP were obtained from ArnetMiner,\footnote{\url{https://aminer.org}} a service that has released multiple versions of this data with the latest papers since $2010$~\citep{tang2008arnetminer,sinha2015overview}.
We chose the latest version (v10) from their collection.

\paragraph{Metadata.} Both datasets feature common metadata, including the year in which the document was published, the number of citations the document has received and the number of references to other documents in the citation network.
A descriptive summary of the complete collection is given in \autoref{tab:descriptive}.
\input{tab-descriptive}

%% file: tab-descriptive.tex
\begin{table}
    \centering
    \begin{tabular}{lll}
    \toprule
    Statistic & Legal opinions & Scientific abstracts\\
    \midrule
    Number of documents & 3,854,738 & 2,408,010 \\
    Years  & 1754--2018 & 1949--2018 \\
    Average number of citations (in-degree) & 7.84 & 39.19 \\
    Average number of references (out-degree) & 7.80 & 9.49 \\
    Length (number of unique word types per document) & 632.22 & 93.10 \\
    \bottomrule
    \end{tabular}
    \caption{Descriptive summary of the two datasets}
    \label{tab:descriptive}
\end{table}

%% file: experiments.tex
\section{Identifying semantic innovations}
We now describe the steps taken to create a list of semantic innovations in these datasets. These innovations are then used to score every document for its progressiveness.
\label{sec:experiments}

\subsection{Preprocessing}
\label{sec:exp-preproc}
For the legal documents, we stripped out HTML and used only the text. The scientific abstracts were available in plain text, but required filtering to identify English-language documents, which we performed using \textit{langid.py}~\citep{lui2012langid}.
In both collections, we converted the text to lowercase before proceeding, and employed \textit{spaCy} for tokenization.\footnote{\url{https://spacy.io/}}

\subsection{Estimating Word Embeddings}
For both document collections, the first (oldest) 500,000 documents were used to learn the early embeddings (matrices $\old{\Embout}$ and $\old{\Embin}$); the most recent 500,000 documents were used to learn the later embeddings (matrices $\new{\Embout}$ and $\new{\Embin}$).
Embeddings were estimated using 
a public tensorflow implementation.\footnote{\url{https://www.tensorflow.org/tutorials/representation/word2vec}, accessed May 2019.}
We ignored tokens with frequency below a predetermined threshold: $5$ for the abstracts and $10$ for the larger dataset of legal opinions.
The maximum size of the context window was set to $10$ tokens.
The number of negative samples was set to $100$. 
The NCE objective was optimized for $50$ epochs and the size of the embeddings for each word was set to $d=300$ dimensions.
While most of the hyperparameters were set to the default values, the size of the embeddings was selected by evaluating on word similarity benchmarks~\citep{luong2013morpho}.

\subsection{Postprocessing}
\label{sec:exp-postproc}
After estimating the embeddings, semantic innovations were identified using the technique described in \nameref{subsec:semantic-innovations}. The number of nearest neighbors used for the computation of the metric was set to $50$.

\paragraph{Names.} In the case of legal opinions, names (e.g., of plaintiffs, defendants, and judges) pose a real difficulty in identifying genuine candidates of semantic innovations.
Although names can be part of semantic innovations (e.g. \example{Nash equilibrium} or \example{Miranda rights}), names often change their distributional statistics due to real-world events rather than semantic change. 
To overcome this problem, we use two heuristics.
We first label a small set of terms if they are names of people, organizations or places, and train a feed-forward neural network to map the embeddings of each word to the label.
This method identifies terms that are distributionally similar to terms that are labeled as names.
Second, we tag a randomly-selected 10\% of the documents for their part of speech and obtain a distribution over parts-of-speech for each vocabulary item, using the pre-trained tagger provided by \example{spaCy}.\footnote{We used \example{spaCy} version 2.0.16 from \url{https://spacy.io/api/tagger}, accessed May 2019. The tagger was trained on the OntoNotes 5 component of the Penn Treebank.}
If a term is either (a) labelled as a name using the first heuristic or, (b) tagged as a proper noun more than 90\% of the time, then it is likely to be a name and is therefore discarded from the candidates of semantic innovations.

\paragraph{Abbreviations.} In the dataset of scientific abstracts, the mention of names is rare, but abbreviations pose a similar challenge. 
We identify abbreviations using a similar heuristic procedure as described above: a term was judged as a likely abbreviation if it was used in all capital (majuscule) letters at least 90\% of the time.
However, as abbreviations can transition to the status of more typical words (e.g., \example{laser}), we chose to discard only those abbreviations which appear fewer than $25$ times in both the early and the later set of abstracts. The abbreviations are common in the scientific abstracts and tend to be dominant as the top ranked semantic changes. For this reason, we kept a higher frequency threshold of $25$ for them to balance between meaningful and spurious changes. 

After applying all the steps mentioned above, we inspected the top words for both legal opinions and computer science abstracts and manually removed names and abbreviations that were not caught by these heuristics, as well as tokenization errors. For each dataset, we retain a list of the $1000$ terms that underwent the most substantial semantic changes, as measured by overlap in their semantic neighborhood (described above). 
Words outside this list have similar embeddings over time; as a result, they are unlikely to yield large progressiveness scores for any documents, and will therefore not impact the overall results. 
As a robustness check, we also performed the regressions using the unfiltered list, and this did not qualitatively change the regression results described below.

%% file: results.tex
\section{Innovations and Innovators}
\label{sec:results}

\paragraph{Semantic changes.}
A few prominent semantic innovations are listed in \autoref{tab:semantic-innovations}.
The innovations in the legal opinions corpus we discover span multiple domains, including financial 
(e.g. \example{laundering}, which earlier exclusively meant washing), socio-political
(e.g. \example{noncitizens}, which was earlier close to \example{tribals} or \example{indians} but has now moved closer in meaning to \example{immigrants}), 
medical (e.g. \example{fertilization}, which was first used in the context of agriculture, but now increasingly refers to human reproduction) 
and technological (e.g. \example{web}, which now refers almost exclusively to the internet).
Our analysis also independently discovers semantic changes in words like \example{cellular} and \example{asylum}, which have previously been identified as semantic changes in other corpora~\citep{kulkarni2015statistically, hamilton2016diachronic, hamilton2016cultural}.

\input{tab-innovations}

In the scientific domain, a common source of semantic innovation is through the use of abbreviations (recall that the filtering steps in the previous section exclude only rare abbreviations).
Examples include 
\example{nfc}, which earlier meant ``neuro-fuzzy controllers'' but lately refers to ``near-field communication'';
\example{ux}, which was used as a short form for unix, but is now increasingly used to mean ``user experience''; 
and 
\example{ssd}, which popularly stood for ``sum-of-squared difference'', but of late additionally means ``solid state drives.''
Another common source of semantic innovations is the creative naming of technological components.
Examples include \example{cloud}, which now refers to services offered through the internet in comparison to its mainstream meaning; \example{spark}, which was earlier popularly used to mean ignition, but has lately been referred to the popular MapReduce framework; and \example{android}, which referred to robots with human appearances, but now commonly refers to the mobile computing operating system. 



\begin{figure}
\centering
\includegraphics[width=\linewidth]{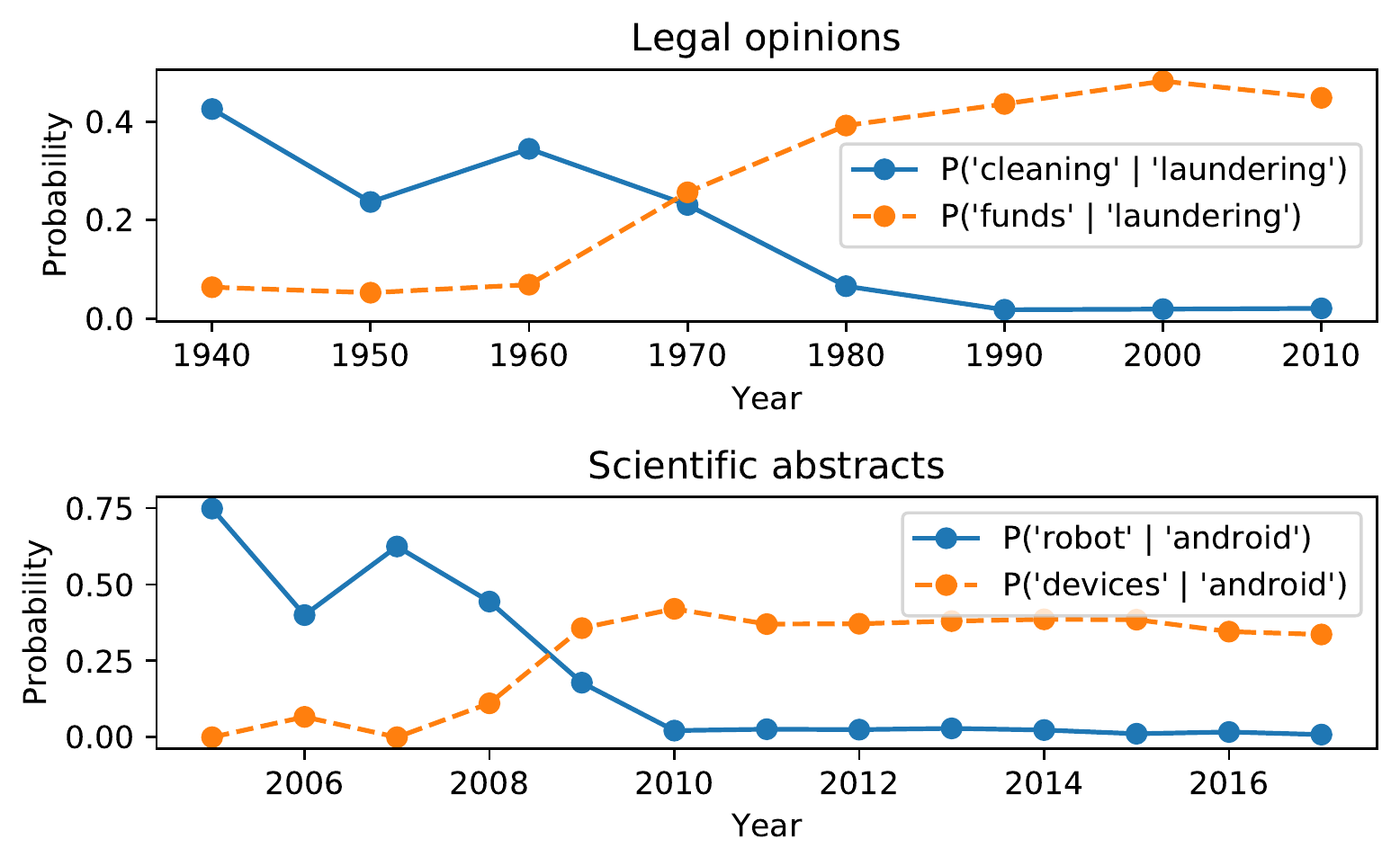}
\caption{Examples of semantic changes identified by the method. In the upper time series, the meaning of the term \example{laundering} evolves to include money laundering, shown here as the increase in the conditional probability of seeing the term \example{laundering} given that \example{funds} appears in a document, in contrast to the conditional probability of \example{laundering} given that \example{cleaning} appears. For this change, a prominent leading document is the opinion in U.S. v Rauhoff (1975). In the lower time series, the meaning of \example{android} evolves to include the mobile phone operating system. A prominent leader of this change is \cite{shabtai2010google}.}
\label{fig:semantic-innovations-timeseries}
\end{figure}

\paragraph{Leading documents.}
Two examples of legal opinions at the leading edge of change according to our metrics are Planned Parenthood vs Casey (505 U.S. 833) and United States v. Talmadge G. Rauhoff, (7th Cir. 1975). 
The landmark 1992 opinion in Planned Parenthood vs Casey was identified by our method as leading a change 
with several semantically progressive terms like \example{fertilization}, \example{provider}, and \example{viability} mentioned in the document. The term \example{fertilization} had previously been used in the context of agriculture, but this decision was an early example of an increasingly common usage in connection with reproductive rights:
\begin{itemize}
\item \example{\ldots two-week gestational increments from \textbf{fertilization} to full term \ldots}
\item \example{\ldots before she uses a post-\textbf{fertilization} contraceptive.}
\end{itemize}
Similarly, the United States v. Talmadge G. Rauhoff, (7th Cir. 1975) scores highly on our measure 
and was one of the first to use \example{laundering} to refer to illegal transfer of money:
\begin{itemize}
\item \example{\ldots \$15,000 as part of the \textbf{`laundering'} process \ldots}
\item \example{\ldots first step in the successful \textbf{laundering} of the funds\ldots}
\end{itemize}
The first mention of the term was quoted, which may indicate a metaphorical intent.

In the scientific domain, the seminal paper on the Android operating system is rated as a semantically progressive document~\citep{shabtai2010google}. 
At that time, the conventional meaning of the term \example{android} was an interactive robot (e.g. \example{\ldots interaction using an \textbf{android} that has human-like appearance\ldots}), but \citeauthor{shabtai2010google} used the now-prevalent meaning as a mobile operating system (e.g. \example{\ldots the \textbf{android} framework \ldots}).
~\autoref{fig:semantic-innovations-timeseries} shows the evolution of the semantic innovations which approximately aligns with the leading documents that our method discovered.

\paragraph{Computational requirements} Our method to score the progressiveness of documents can effectively leverage computational resources without putting a heavy burden on them. Following~\autoref{eq:progressiveness}, we calculate the normalization terms for a specific change just once for the entire corpus to speed up the calculation considerably. Moreover, the progressiveness calculation is also embarrassingly parallelizable; that is, scores for different documents can be computed in parallel. We calculated the progressiveness scores of millions of documents on an 80 processor machine with 256GB memory in just over a couple of hours.  


%% file: tab-innovations.tex
\begin{table*}
\small
\centering
\begin{tabular}{@{}lllp{1.1in}p{1.8in}@{}}
\toprule
Doc. type & Innovations & Old usage & New usage & Example document with new usage \\  
\midrule
Legal & \example{laundering} & \example{\textbf{laundering} clothes}& \example{\textbf{laundering} funds} & United States v. Talmadge G. Rauhoff, (7th Cir. 1975)\\
 & \example{asylum}& \example{insane \textbf{asylum}}& \example{political \textbf{asylum}}& Bertrand v. Sava, (S.D.N.Y. 1982)\\
& \example{fertilization}& \example{soil \textbf{fertilization}} & \example{post-\textbf{fertilization} contraceptive} & Planned Parenthood vs Casey (505 U.S. 833)\\[4ex]

Science & \example{ux} & \example{hp-\textbf{ux}} & \example{user experience (\textbf{ux})}& \cite{hassenzahl2006user} \\
& \example{surf} & \example{\textbf{surf} the internet}& \example{descriptor \textbf{surf}} & \cite{bay2008speeded} \\
& \example{android} & \example{intelligent \textbf{android}}& \example{google's \textbf{android}} & \cite{shabtai2010google}\\
\bottomrule
\end{tabular}

\caption{Examples of semantic innovations identified by our method for both the datasets.}.
\label{tab:semantic-innovations}
\end{table*}

%% file: testing.tex
\section{Innovation and Influence}
\label{sec:testing}
While the examples in the previous section are suggestive of the validity of our method for identifying innovations and innovators, additional validation is necessary. Lacking large-scale manual annotations for the semantic progressiveness of legal opinions or scientific abstracts, we instead measure \emph{influence}, as quantified by citations. Specifically, we investigate the hypothesis that more citations will accrue to documents that our metrics judge to be semantically progressive. 

Note that we do not hypothesize a one-way causal relationship from semantic innovation to influence. Semantic progressiveness may cause some documents to be highly cited, due to the introduction or usage of helpful new terminology. But it also seems likely that documents that are well-cited for more fundamental reasons --- e.g., significant methodological innovations in science, foundational precedents in law --- will also exert an outsize effect on language. For example, the highly cited paper on Latent Dirichlet Allocation~\citep{blei2003latent} introduced a new meaning for the term \example{LDA} (which also refers to linear discriminant analysis), but in this case it is likely that the underlying cause is the power of the method rather than the perspicacity of the name. The key point of these evaluations is to test the existence of a previously unknown correlation between language and citation networks, and to provide a further validation of our measure of semantic progressiveness.

\begin{figure}
    \centering
    \includegraphics[width=\linewidth]{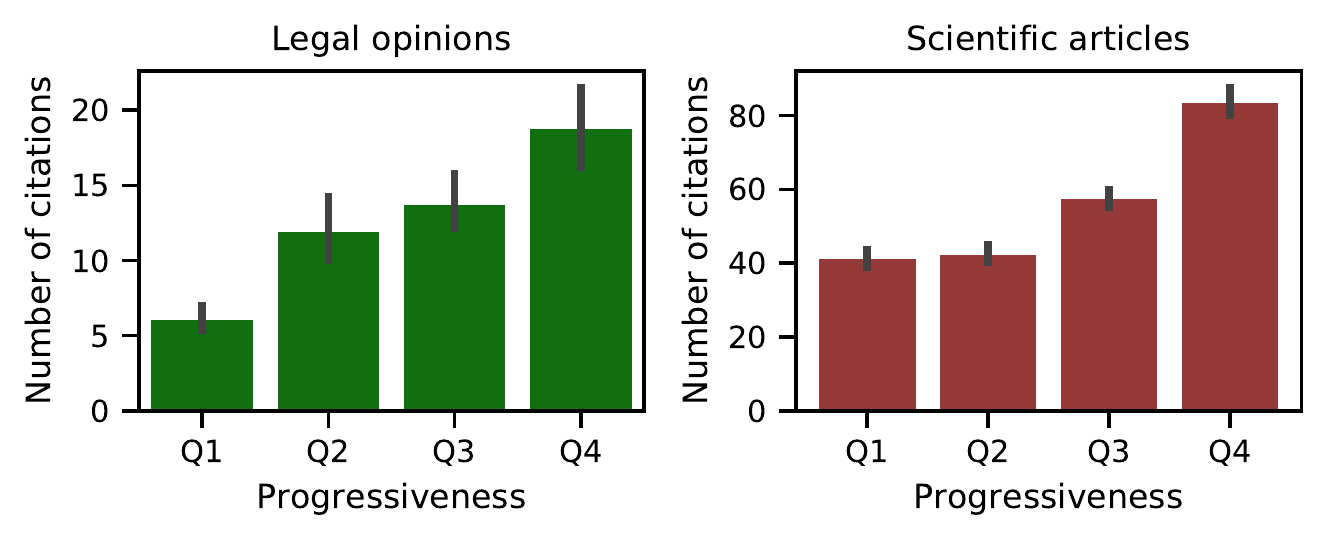}
    \caption{The univariate relationship between the number of citations and our measure of semantic progressiveness. For both the legal opinions and the scientific articles, the citations increase for more progressive documents.}
    \label{fig:cites-vs-progressiveness}
\end{figure}

\subsection{Univariate analysis}
\autoref{fig:cites-vs-progressiveness} shows the number of citations for each quartile of our progressiveness measure, indicating a steady increase in both datasets.
This figure excludes documents that do not include any of the terms identified as having changing semantics. 
We also exclude documents predating 1980, which skew the population with a few landmark examples with vast citation counts; these documents are included in the multivariate analysis that follows. 

\subsection{Multivariate analysis}
There are many factors that drive citation counts, such as age, length, and content~\citep{fowler2007network, van2012citation}.
Some of these factors may be correlated with semantic progressiveness, confounding the analysis: for example, older documents have more chances to be cited, but are unlikely to lead a semantic change that would be captured by our metrics. 
To control for these additional predictors, we formulate the problem as a multivariate regression. The dependent variable is the number of citations, and the predictors include our measure of semantic progressiveness, as well as a set of controls. 
As the number of citations is a count variable, we fit a Poisson regression model.\footnote{In cases of overdispersion (high variance), negative binomial regression is preferred to Poisson regression~\citep{greene2003econometric}. However, the Cameron-Trivedi test~\citep{cameron1990regression} did not detect overdispersion in our data.}

\input{tab-abstracts}

\subsubsection{Regression models}
To assess the relevance of semantic progressiveness, we compare against two baseline models, which include covariates that capture structural information about each document: the number of outgoing references that a document makes; its age; its length, operationalized as the number of unique types; and the number of authors for the document (available only for scientific articles). 
The baseline also incorporates a lightweight model of document content, to account for the fact that some topics may get cited more than others. Specifically, we fit a bag-of-words regression model on a small subset of documents~\cite[similar to][]{yogatama-EtAl:2011:EMNLP}, and use its prediction as a covariate in the multivariate regression. We refer to this covariate as BoWs. This baseline is referred to as M1.

The second baseline, M2, includes all the covariates from M1, and an additional covariate for the number of unique semantic innovations present in the document. This is aimed to tease out the effect of the \emph{presence} of words with changing semantics from the extent to which the document employs the more contemporary meaning, as captured by our measure of semantic progressiveness. We refer to this covariate as \# Innovs. 

To test the effect of semantic progressiveness, we create two experimental models, M3 and M4, which use the
$z$-scores described earlier. 
In M3, the $z$-score is included as a raw value; in M4 it is binned into quartiles. Note that for M4, the bottom quartile (Q1) receives a coefficient of zero by default, so that the model is not underdetermined.

We compare these models by goodness-of-fit, which is a standard technique from quantitative social science~\citep{greene2003econometric}. We compute the log-likelihood for each model; under the null hypothesis that the more complex model is no better than the baseline, the log-likelihood ratio has a $\chi^2$ distribution with degrees of freedom equal to the number of parameters in the more expressive model. If the observed log-likelihood ratio is unlikely to arise under this distribution, then we can reject the null hypothesis. This approach is similar in spirit to the Akaike Information Criterion (AIC), which also penalizes the log-likelihood by the number of parameters. We can also measure the effect size by examining the regression coefficients: the value of each coefficient corresponds to an increase in the log of the expected number of citations under the Poisson model.

\subsubsection{Results}
The regressions reveal a strong relationship between semantic progressiveness and citation count.
For the scientific abstracts (\autoref{tab:regress-abstracts}), M3 and M4 obtain a significantly better fit than M1 ($\chi^2(2) = 137767, p \approx 0$ and $\chi^2(4) = 250479, p \approx 0$ respectively). M3 and M4 also obtain a significantly better fit than M2 ($\chi^2(1) = 130176, p \approx 0$ and $\chi^2(3) = 242889, p \approx 0$ respectively). The effect sizes are relatively large: the coefficient of $0.698$ for top quartile of semantic progressiveness corresponds to an increase in the expected number of citations by a factor of $2$, in comparison with documents in the bottom quartile.

\input{tab-legal}

The story is similar for the legal opinions in \autoref{tab:regress-legal}, with only minor differences. Both M3 and M4 significantly improve the goodness of fit over the baseline M1 ($\chi^2(2) = 8352,p \approx 0$ and $\chi^2(4) = 7164$ respectively) and the baseline M2 ($\chi^2(1) = 3758,p \approx 0$ and $\chi^2(3) = 2571$ respectively), indicating again that semantic progressiveness of the document is highly predictive of the number of incoming citations, even after controlling for several covariates. The coefficient of $0.47$ for the top quartile of progressiveness corresponds to an increase in the expected number of citations by a factor of $1.6$, as compared to the bottom quartile. 

Since age is a strong predictor of citation count for both the datasets, we also compared the distribution of publication years in each quartile to rule out the possibility that the higher quartiles could be acting as another proxy for age. Using a Kolmogorov Smirnoff test, we tested whether the distributions in each pair of quartiles were different, with the null hypothesis being that they were statistically equivalent. We found that statistical equivalence could not be ruled out (on scientific abstract collection the p-values ranged from $0.3$ to $0.92$ and on the courts opinion collection the p-values ranged from $0.1$ to $0.18$) meaning that the semantic progressiveness scoring scheme does not discriminately favor old or recent documents. Overall, these results indicate that our measure of semantic progressiveness adds substantial new information to the array of covariates included in the baseline models, and that semantically progressive documents receive significantly more citations.



%% file: tab-abstracts.tex
\begin{table}[!htbp]

\centering

\small
\begingroup
\setlength{\tabcolsep}{3pt} 
\input{results-dblp}

\endgroup
\caption{Poisson regression analysis of citations to scientific abstracts. Each column indicates a model, each row indicates a predictor, and each cell contains the coefficient and, in parentheses, its standard error. Log likelihood is in millions of nats.}

\label{tab:regress-abstracts}

\end{table}

%% file: results-dblp.tex
\begin{tabular}{p{2.5cm}p{1.4cm}p{1.4cm}p{1.4cm}p{1.4cm}}
\toprule

Predictors & M1 & M2 & M3 & M4\\

\midrule

Constant&1.983&1.943&2.032&1.770\\
 &(0.001)&(0.001)&(0.001)&(0.001)\\[4pt]
Outdegree&0.009&0.009&0.009&0.009\\
 &(0.000)&(0.000)&(0.000)&(0.000)\\[4pt]
\# Authors&0.055&0.054&0.054&0.054\\
 &(0.000)&(0.000)&(0.000)&(0.000)\\[4pt]
Age&0.079&0.079&0.078&0.073\\
 &(0.000)&(0.000)&(0.000)&(0.000)\\[4pt]
Length&0.002&0.002&0.002&0.002\\
 &(0.000)&(0.000)&(0.000)&(0.000)\\[4pt]
BoWs&0.000&0.000&0.000&0.000\\
 &(0.000)&(0.000)&(0.000)&(0.000)\\[4pt]
\# Innovs& &0.028&-0.010&-0.034\\
 & &(0.000)&(0.000)&(0.000)\\[4pt]
Prog.& & &0.137& \\
 & & &(0.000)& \\[4pt]
Prog. Q2& & & &0.179\\
 & & & &(0.001)\\[4pt]
Prog. Q3& & & &0.431\\
 & & & &(0.001)\\[4pt]
Prog. Q4& & & &0.698\\
 & & & &(0.001)\\[4pt]
Log Lik.&-13.07&-13.06&-12.93&\textbf{-12.82}
\\
\bottomrule

\end{tabular}

%% file: tab-legal.tex
\begin{table}[!htbp]

\centering
\small

\begingroup
\setlength{\tabcolsep}{6pt} 
\input{results-ops}

\endgroup
\caption{Poisson regression analysis of citations to legal documents. Each column indicates a model, each row indicates a predictor, and each cell contains the coefficient and, in parentheses, its standard error.
}
\label{tab:regress-legal}

\end{table}

%% file: results-ops.tex
\begin{tabular}{lp{1.4cm}p{1.4cm}p{1.4cm}p{1.4cm}}
\toprule

Predictors & M1 & M2 & M3 & M4\\

\midrule

Constant&1.614&1.421&1.476&1.168\\
 &(0.003)&(0.004)&(0.004)&(0.006)\\[4pt]
Outdegree&0.022&0.020&0.021&0.020\\
 &(0.000)&(0.000)&(0.000)&(0.000)\\[4pt]
Age&0.009&0.011&0.010&0.010\\
 &(0.000)&(0.000)&(0.000)&(0.000)\\[4pt]
Length&0.000&-0.000&-0.000&-0.000\\
 &(0.000)&(0.000)&(0.000)&(0.000)\\[4pt]
BoWs&-0.000&-0.000&-0.000&-0.000\\
 &(0.000)&(0.000)&(0.000)&(0.000)\\[4pt]
\# Innovs& &0.054&0.045&0.042\\
 & &(0.001)&(0.001)&(0.001)\\[4pt]
Prog.& & &0.094& \\
 & & &(0.001)& \\[4pt]
Prog. Q2& & & &0.384\\
 & & & &(0.007)\\[4pt]
Prog. Q3& & & &0.382\\
 & & & &(0.007)\\[4pt]
Prog. Q4& & & &0.470\\
 & & & &(0.007)\\[4pt]
Log Lik.&-415195&-410601&\textbf{-406843}&-408031
\\
\bottomrule

\end{tabular}

%% file: related.tex
\section{Related Work}
\label{sec:related}
\subsection{Language change}
Language change has been a topic of great general interest. Much of the early computational work on language change focused on tracking the \emph{frequency} of lexical items, rather than their meaning. \cite{michel2011quantitative} track changes in word frequency in large books corpora, and link these changes to social-cultural trends and events. \cite{danescu2013no} compare the adoption rates of words between community members. 
\cite{eisenstein2014diffusion} track the diffusion of new words over geographical regions, and \cite{goel2016social} model diffusion across social networks.
Measures of linguistic progressiveness in this line of work are also based on frequency and other dependent statistics, such as cross entropy~\citep{danescu2013no} or \example{tf-idf}~\citep{kelly2018measuring}.

More recently, several methods have been proposed to learn diachronic word embeddings as a means to track language change at a finer semantic level. These methods include matrix decomposition~\citep[e.g.,][]{yao2018dynamic}, Bayesian inference~\citep[e.g.,][]{wijaya2011understanding,frermann2016bayesian,bamler2017dynamic}, and neural word embeddings~\citep[e.g.,][]{kim2014temporal,kulkarni2015statistically,hamilton2016diachronic,rosenfeld2018deep}.
Diachronic word embeddings have shown success in identifying linguistic~\citep{hamilton2016cultural} and sociocultural changes over time~\citep{garg2018word}. Two surveys review the existing research on diachronic language change through word embeddings~\citep{kutuzov2018diachronic,tahmasebi2018survey}. Contextual word representations~\citep[e.g.,][]{devlin2019bert}, which have produced state-of-the-art results in natural language processing, have also been utilized to identify semantic changes~\citep{giulianelli2020analysing}. However, despite these successes, prior work has not provided methods to identify the documents at the forefront of semantic change. Our work specifically addresses this gap.

Another body of work has used topic modeling to study changes over time~\citep[e.g.,][]{blei2006dynamic,wang2006topics,mimno2012computational}.
Of particular relevance is the use of topical changes in scientific literature to discover documents with the most scholarly impact~\citep{gerrish2010language,hall2008studying}.
We argue that these approaches are complementary.
While topic models provide a macro-level view of the concerns and interests of a set of writers, word embeddings provide a more fine-grained perspective by demonstrating shifts in meaning of individual terms.
Topic models are centered at the document level, and so make it easy to identify innovators; our work extends this capability to embedding-based analysis of semantic change.
\subsection{Citation impact}
The number of citations a document receives has long been used as a proxy for the impact and influence of scientific articles~\citep{fortunato2018science}, legal opinions~\citep{fowler2007network}, as well as researchers and scientific trends~\citep{borner2004simultaneous}. 
Dynamic models capturing the mechanics of attention have been modestly successful in predicting long-term scientific impact~\citep{barabasi13}. Other models accounting for changing language have been used to identify important new topics~\citep{borner2004simultaneous} or to estimate the influence of papers on one another~\citep{dietz2007unsupervised}. 
In a different domain, progressiveness as measured in terms of textual dissimilarity with past patents and textual similarity with future patents is shown to be predictive of future citations of a patent~\citep{kelly2018measuring}. Our quantitative insights in this work are similar to that of \cite{kelly2018measuring} but our measure of language change is more fine-grained and is based on semantic changes, instead of textual similarity which can depend on topics or events. 

%% file: discussion.tex
\section{Conclusion}
\label{sec:discussion}
This paper shows how to identify the leading examples of semantic change, by leveraging the models underlying diachronic word embeddings. This enables us to test the hypothesis that semantically progressive documents --- that is, documents that use words in ways that reflect a change in progress --- tend to receive more citations. This technique has potential applicability in the digital humanities, computational social science, and scientometrics~\cite[the study of science itself; see][]{van1997scientometrics}. Our current method of identifying semantically progressive usage is limited to words. For future work, we are interested in extending the current method beyond words to phrases and grouping phrases and words with semantic similarity together -- for example, this would enable the word \example{LDA} and the phrase \example{Latent Dirichlet Allocation} to have similar embeddings and potentially improve the results for progressive usage identification. In future work, we are also interested to assess how semantically progressive documents are received by their audiences, and to explore semantic change as a site of linguistic contestation. For example, recent work has linked diachronic word embeddings to gender and ethnic stereotypes in large-scale datasets of books~\citep{garg2018word}. Our method could link author and audience covariates with the documents that led and trailed changes in these stereotypical associations, providing new insight on these historical trends. Finally, contemporaneous work has demonstrated the use of contextualized embeddings, highly powerful word representations, in detecting semantic changes~\citep{giulianelli2020analysing}. In future, we are interested to extend contextual embeddings to measure semantic progressiveness. 


%% file: supplementary.tex
\appendix
\section{Robustness Checks}
We conducted a series of stability and robustness checks to verify that our proposed method is reliable. Learning word embeddings using NCE or similar such methods is prone to stability issues, in particular due to random initialization~\citep{antoniak2018evaluating,burdick2018factors}. We ran our pipeline of learning word embeddings, identifying semantic innovations, and measuring the semantic progressiveness of documents for different random initialization.

\subsection{Word Embeddings Stability}
Since our proposed calculation of progressiveness of every document relies heavily on the word embeddings, high variance in the word embeddings due to random initialization can potentially affect the calculation. We tested the performance of word embeddings under different initialization on benchmark testsets to verify that our method is quite stable. Specifically, we evaluate the quality of the word embeddings on analogy and word similarity tasks for three runs, each differing in the initialization point. The results are in~\autoref{tab:benchmark-analogy} and~\autoref{tab:benchmark-wordsim} respectively. 
\input{tab-benchmarks}

\subsection{Semantic Innovations Stability}
Even though the performance on extrinsic benchmarks points to word embeddings being of similar quality irrespective of random initialization, it does not necessarily mean that there is low variance in uncovering semantic changes. To make this explicit, we show the top 10 top semantic changes identified for each run on the CourtListener text collection in~\autoref{tab:top-innovs-cl} and for the DBLP collection in~\autoref{tab:top-innovs-dblp}.
\input{tab-top-innovations}

\subsection{Robust Semantic Progressiveness}
The word embeddings and the discovered semantic innovations are stable despite differences in initialization. But the embeddings and the semantic innovations are dependent variables in our calculation of the semantic progressiveness. We performed another quantitative check to show that the progressiveness scores of documents are correlated across different runs. ~\autoref{tab:semcorr} shows the spearman rank correlation across random pairs of runs for both the text collections. As can be seen the spearman rank correlation is extremely high, meaning that even the small amount of noise that is added to the embeddings due to initialization is canceled through the calculation of progressiveness scores.
\input{tab-spearmanr}

\section{Alternative measurement of semantic progressiveness}
The results from the multivariate regressions using an alternative measure of semantic progressiveness. In this scoring scheme the progressiveness per document is calculated as the number of innovations in the document for which the progressiveness score is greater than the median progressiveness score across all semantic innovations. \autoref{tab:regress-abstracts-numinnovs} contains the results for the DBLP collection and the \autoref{tab:regress-legal-numinnovs} contains the results for the collection of court opinions. 
\input{tab-abstracts-numinnovs}
\input{tab-legal-numinnovs}

%% file: tab-benchmarks.tex
\begin{table}
    \centering
    \begin{tabular}{p{1cm}p{1cm}p{1cm}p{1cm}p{1cm}}
    \toprule
    Runs & \multicolumn{2}{c}{CourtListener} & \multicolumn{2}{c}{DBLP} \\
    \midrule
    & Early & Later & Early & Later \\
    \midrule
    1 & 20.5 & 22.7 & 11.8 & 16.6 \\
    2 & 20.9 & 22.7 & 11.5 & 16.5 \\
    3 & 20.9 & 22.4 & 11.8 & 16.6 \\
    \bottomrule
    \end{tabular}
    \caption{Accuracy (in \%) of word embeddings on the analogy testset~\citep{mikolov2013efficient}}
    \label{tab:benchmark-analogy}
\end{table}

\begin{table}
    \centering
    \begin{tabular}{p{1cm}p{1cm}p{1cm}p{1cm}p{1cm}}
    \toprule
    Runs & \multicolumn{2}{c}{CourtListener} & \multicolumn{2}{c}{DBLP} \\
    \midrule
    & Early & Later & Early & Later \\
    \midrule
    1 & 0.42 & 0.42 & 0.35 & 0.45 \\
    2 & 0.42 & 0.43 & 0.34 & 0.45 \\
    3 & 0.42 & 0.43 & 0.34 & 0.44 \\
    \bottomrule
    \end{tabular}
    \caption{Spearman correlation of word embeddings on the word similarity testset~\citep{bruni2012distributional}}
    \label{tab:benchmark-wordsim}
\end{table}

%% file: tab-top-innovations.tex
\begin{table}
    \centering
    \begin{tabular}{c|p{10cm}}
    \toprule
    Runs & Top Semantic Innovations\\
    \midrule
    1 & \textbf{\example{underpinned}}, \textbf{\example{lodgment}}, \example{recomissioned}, \example{disentangling}, \textbf{\example{entrenchment}}, \example{forensically}, \example{replications}, \textbf{\example{fringe}}, \textbf{\example{bonded}}, \textbf{\example{clout}} \\
    2 & \textbf{\example{entrenchment}}, \example{cloaks}, \textbf{\example{underpinned}}, \example{replications}, \example{unshackled}, \textbf{\example{lodgment}}, \example{origination}, \textbf{\example{clout}}, \textbf{\example{bonded}}, \textbf{\example{fringe}} \\
    3 & \textbf{\example{underpinned}}, \textbf{\example{lodgment}}, \textbf{\example{entrenchment}}, \example{cloaks}, \example{forensically}, \example{origination}, \textbf{\example{clout}}, \example{telegraphing}, \textbf{\example{fringe}}, \textbf{\example{bonded}} \\
    \bottomrule
    \end{tabular}
    \caption{Top semantic changes across different runs for Courtlistener text collection. Six changes appear in the top ten across all three runs, as shown in bold.}
    \label{tab:top-innovs-cl}
\end{table}

\newcommand{\bexample}[1]{\textbf{\example{#1}}}
\begin{table}
    \centering
    \begin{tabular}{c|p{10cm}}
    \toprule
    Runs & Top Semantic Innovations\\
    \midrule
    1 & \textbf{\example{osn}}, \bexample{ux}, \bexample{asd}, \bexample{ros}, \bexample{ble}, \bexample{mtc}, \bexample{hesitant}, \example{apps}, \bexample{nfc}, \bexample{app} \\
    2 & \bexample{ux}, \bexample{ble}, \textbf{\example{osn}}, \bexample{asd}, \bexample{app}, \bexample{hesitant}, \bexample{mtc}, \example{ppi}, \bexample{nfc}, \bexample{ros} \\
    3 & \textbf{\example{osn}}, \bexample{ux}, \bexample{ros}, \bexample{hesitant}, \bexample{ble}, \example{ppi}, \bexample{asd}, \bexample{app}, \bexample{mtc}, \bexample{nfc} \\
    \bottomrule
    \end{tabular}
    \caption{Top semantic changes across different runs for DBLP text collection. Nine changes appear in the top ten across all three runs, as shown in bold.}
    \label{tab:top-innovs-dblp}
\end{table}

%% file: tab-spearmanr.tex
\begin{table}
    \centering
    \begin{tabular}{c|c|c}
    \toprule
    Runs & Scientific abstracts & Court opinions\\
    \midrule
    1-2 & 0.994 & 0.995\\
    2-3 & 0.996 & 0.993\\
    1-3 & 0.992 & 0.995\\
    \bottomrule
    \end{tabular}
    \caption{Spearman rank correlation across random pairs of runs for both the text collections.}
    \label{tab:semcorr}
\end{table}

%% file: tab-abstracts-numinnovs.tex
\begin{table}[!htbp]

\centering

\small
\begingroup
\setlength{\tabcolsep}{3pt} 
\input{results-dblp-numinnovs}
\endgroup
\caption{Poisson regression analysis of citations to scientific abstracts. Each column indicates a model, each row indicates a predictor, and each cell contains the coefficient and, in parentheses, its standard error. Log likelihood is in millions of nats.}

\label{tab:regress-abstracts-numinnovs}

\end{table}

%% file: results-dblp-numinnovs.tex
\begin{tabular}{@{}lp{1.3cm}p{1.3cm}p{1.3cm}@{}}
\toprule


  & M1 & M2 & M3\\
\midrule

Constant&2.078&2.011&2.008\\
 &(0.001)&(0.001)&(0.001)\\[4pt]
Outdegree&0.010&0.010&0.010\\
 &(0.000)&(0.000)&(0.000)\\[4pt]
\# Authors&0.024&0.024&0.024\\
 &(0.000)&(0.000)&(0.000)\\[4pt]
Age&0.074&0.077&0.076\\
 &(0.000)&(0.000)&(0.000)\\[4pt]
Length&0.002&0.002&0.002\\
 &(0.000)&(0.000)&(0.000)\\[4pt]
BoWs&0.000&0.000&0.000\\
 &(0.000)&(0.000)&(0.000)\\[4pt]
Prog.& &0.049& \\
 & &(0.000)& \\[4pt]
Prog. Q2& & &0.105\\
 & & & (0.001)\\[4pt]
Prog. Q3& & &0.045\\
 & & & (0.001)\\[4pt]
Prog. Q4& & &0.137\\
 & & & (0.001)\\[4pt]
Log Lik.&-12.923&\textbf{-12.891}&-12.912
\\
\bottomrule
\end{tabular}

%% file: tab-legal-numinnovs.tex
\begin{table}[!htbp]

\centering
\small

\begingroup
\setlength{\tabcolsep}{6pt} 
\input{results-ops-numinnovs}
\endgroup
\caption{Poisson regression analysis of citations to legal documents. Each column indicates a model, each row indicates a predictor, and each cell contains the coefficient and, in parentheses, its standard error.
}
\label{tab:regress-legal-numinnovs}

\end{table}

%% file: results-ops-numinnovs.tex
\begin{tabular}{@{}lp{1.4cm}p{1.4cm}p{1.4cm}@{}}
\toprule

  & M1 & M2 & M3 \\

\midrule

Constant&1.612&1.515&0.963\\
 &(0.003)&(0.004)&(0.007)\\[4pt]
Outdegree&0.019&0.020&0.019\\
 &(0.000)&(0.000)&(0.000)\\[4pt]
Age&0.011&0.012&0.017\\
 &(0.000)&(0.000)&(0.000)\\[4pt]
Length&0.000&0.000&0.000\\
 &(0.000)&(0.000)&(0.000)\\[4pt]
BoWs&0.000&0.000&0.000\\
 &(0.000)&(0.000)&(0.000)\\[4pt]
Prog.& &0.051& \\
 & &(0.001)& \\[4pt]
Prog. Q2& & &0.577\\
 & & &(0.006)\\[4pt]
Prog. Q3& & & 0.615\\
 & & & (0.007)\\[4pt]
Prog. Q4& & &0.745\\
 & & &(0.008)\\[4pt]
Log Lik.&-429096&-427724&\textbf{-423474}
\\
\bottomrule

\end{tabular}